\documentclass[letterpaper, 10 pt, conference]{ieeeconf}  

\IEEEoverridecommandlockouts                              

\usepackage{graphicx}
\usepackage[inkscapelatex=false]{svg}
\usepackage{booktabs}
\usepackage{tabularx}
\usepackage{array}
\usepackage{url}
\usepackage{fancyhdr}

\title{\LARGE \bf
Autoware in Construction: Gap Analysis and LiDAR Perception Toward Off-Road Autonomous Driving \
}

\author{Yu Otsuki$^{1}$, Teja Emmey$^{2}$, Sena Matsushita$^{2}$, Akiro Harada$^{2}$, and Takeshi Miura$^{2}$
\thanks{This work was conducted while the first author was with TIER IV.}
\thanks{$^{1}$The author is an Assistant Professor at the Department of Civil and Environmental Engineering, University of California, Irvine (UCI), Irvine, CA, USA.
        {\tt\small yu.otsuki@uci.edu}}%
\thanks{$^{2}$The authors are with TIER IV, Inc.,
        Shinagawa, Tokyo, Japan.
        {\tt\small \{teja.emmey, sena.matsushita, akiro.harada, takeshi.miura\}@tier4.jp}}%
}

\begin{document}

\maketitle

\fancypagestyle{firstpage}{
    \fancyhf{}
    \renewcommand{\headrulewidth}{0pt}
    \renewcommand{\footrulewidth}{0pt}
    \fancyfoot[C]{\footnotesize Published in the 2026 IROS 5th Workshop on Future of Construction.}
}

\thispagestyle{firstpage}
\pagestyle{empty}

\begin{abstract}
Autoware is an open-source autonomous driving software platform widely adopted by researchers and industry developers. Originally developed primarily for public-road applications, including passenger vehicles, taxis, and buses, Autoware is increasingly being extended to off-road environments such as construction and agricultural sites. Construction sites, however, differ fundamentally from public roads and challenge many assumptions underlying conventional autonomous driving systems. They are characterized by unstructured terrain, airborne dust, continuously evolving site conditions, and construction-specific objects. This paper presents lessons learned from ongoing efforts to extend an Autoware-based autonomous driving system to large dump trucks operating at construction sites. We identify gaps between public-road and off-road construction applications across the sensing, mapping, localization, perception, planning, and control modules of the Autoware stack. We then discuss potential solutions for addressing these gaps, with particular emphasis on ongoing LiDAR-based perception pipeline development and findings from field testing. Finally, we propose a roadmap toward end-to-end autonomous driving for construction vehicles.
\end{abstract}

\section{INTRODUCTION}

\begin{figure}[thpb]
    \centering
    \includegraphics[width=0.95\columnwidth]{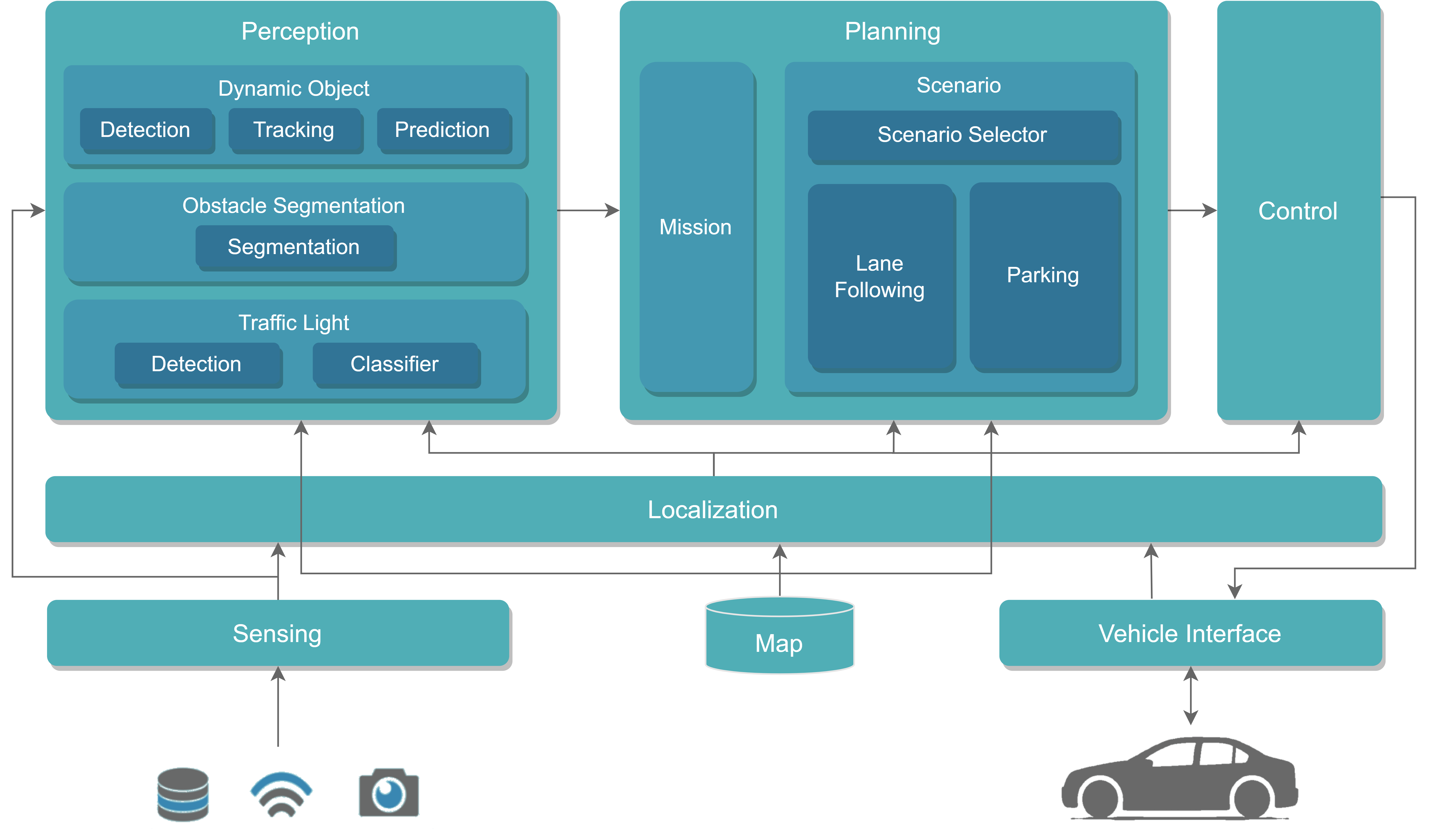}
    \caption{Overview of Autoware modular architecture \cite{autoware_architecture}.}
    \label{fig:autoware_architecture_v1}
\end{figure}

Autoware is a modular, open-source autonomous driving software stack that has supported research and development across a wide range of applications, including passenger vehicles, buses, taxis, delivery vehicles, industrial vehicles, and racing cars. Originally implemented on ROS 1~\cite{kato2018autoware} and since migrated to ROS 2, its architecture integrates sensing, mapping, localization, perception, planning, control, and vehicle-interface components (Fig. \ref{fig:autoware_architecture_v1}). This modularity allows individual components to be replaced or extended for each operational design domain (ODD).

\begin{figure}[thpb]
    \centering
    \includegraphics[width=0.9\columnwidth]{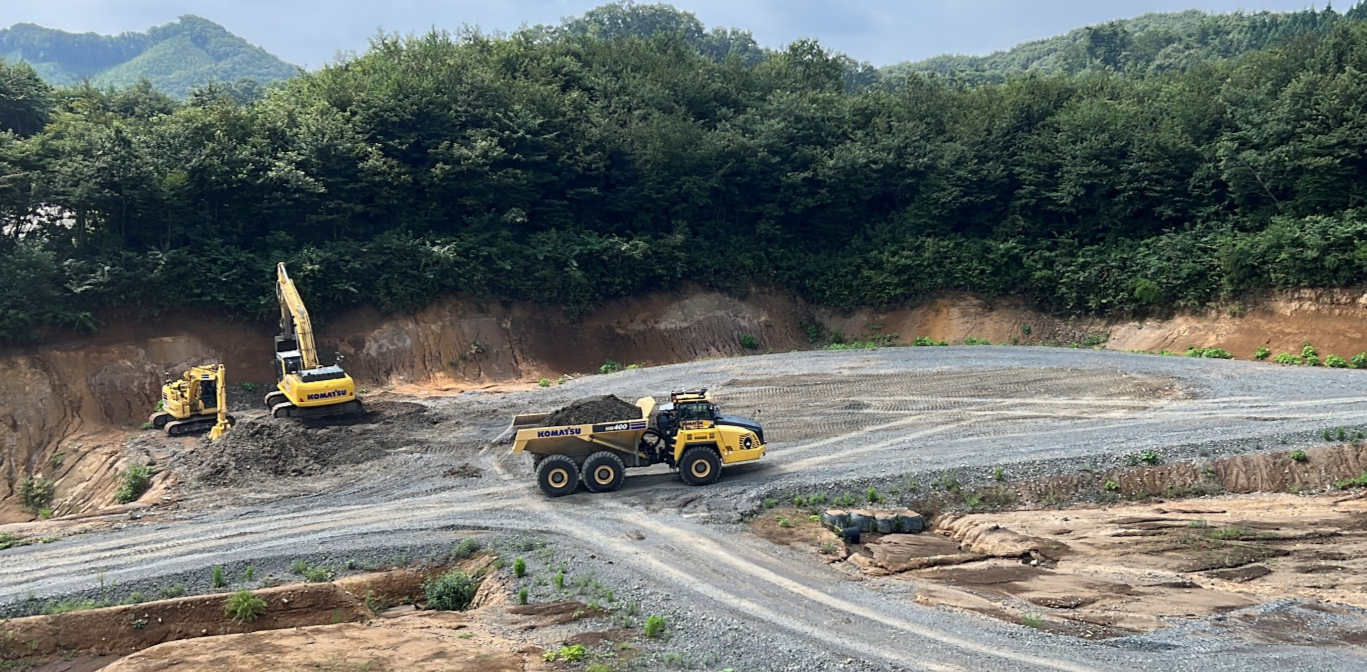}
    \caption{Autoware-integrated autonomous construction dump truck tested in an off-road environment.}
    \label{fig:truck}
\end{figure}

Construction sites represent a substantially different ODD from public roads. Construction environments introduce rough terrain, airborne dust, continuously changing site geometry, feature-poor environments, construction-specific objects, and coordinated operations among heterogeneous machines~\cite{uchiito2025robotics}. These conditions challenge several assumptions embedded in public-road autonomous driving systems and require adaptations throughout the Autoware stack. Recently, the Autoware Foundation has initiated an off-road pilot encompassing mining, racing, and planetary applications~\cite{autoware_offroad}.

As part of our effort to extend Autoware to construction environments, we are integrating the Autoware software stack into a large construction dump truck, as shown in the center of Fig.~\ref{fig:truck}. The vehicle has been retrofitted with LiDAR, camera, GNSS, and IMU sensors, onboard computers, and an electronic control unit interface.

Based on this ongoing development effort, this paper makes three contributions. First, it presents a module-level gap analysis between the existing Autoware architecture, which was developed primarily for public-road autonomy, and the requirements of off-road autonomy in construction environments. Second, it distills lessons learned from the development and field testing of a LiDAR-based perception pipeline for construction sites. Third, it discusses a roadmap toward end-to-end (E2E) autonomous driving for construction vehicles.

\begin{table*}[t]
\caption{Technology gaps in autonomous driving between public roads and construction sites.}
\label{tab:technology_gap}
\centering
\small
\renewcommand{\arraystretch}{1.08}
\setlength{\tabcolsep}{4pt}

\begin{tabularx}{\textwidth}{
    @{}
    >{\raggedright\arraybackslash}p{0.14\textwidth}
    >{\raggedright\arraybackslash}X
    >{\raggedright\arraybackslash}X
    @{}
}
\toprule
\textbf{Modules}
& \textbf{Public-road capabilities available in Autoware}
& \textbf{Off-road construction challenges and requirements} \\
\midrule

Sensing
& Supports LiDAR, camera, radar, GNSS, and IMU sensors. Standard sensor drivers and sensor calibration routines are also provided.
& Dust can cause occlusion. Mud can contaminate sensor surfaces. Large vehicle bodies can create blind spots. IMU measurements are degraded by strong vibrations. Severe vertical vibration requires 3D LiDAR distortion correction.  \\
\midrule

Mapping
& Lane geometry and surrounding structures are static, so high-definition maps can be maintained offline. Environments are geometrically feature-rich, with abundant vertical structures.
& Construction sites evolve continuously through excavation, grading, and material placement, requiring frequent map updates. \\
\midrule

Localization
& GNSS/RTK can be unreliable in tunnels, urban canyons, and beneath elevated roads. Map-based localization performs well with feature-rich high-definition maps. Planar localization in $x$, $y$, and yaw is often sufficient.
& GNSS/RTK is generally reliable under open skies. Map-based localization is challenged by feature-poor, changing environments. Full 6-DOF pose estimation is often required due to slopes and rough terrain. \\
\midrule

Perception
& Supports detection, tracking, and sensor fusion using both non-ML and ML-based approaches. Its ML detectors support objects such as cars, trucks, buses, bicycles, and pedestrians.
& Relevant objects include heavy construction machinery, passenger vehicles, construction workers, cones, rocks, berms, and temporary structures. The drivable space is unstructured and must be identified. \\
\midrule

Planning
& Scenario-based, such as lane driving and parking on predefined road networks, governed by traffic rules and dominated by forward driving.
& Haulage roads are stable over days to weeks and can be given as  waypoints, but loading and dumping areas change continuously and require coordination with excavators and dozers. Traversability must be evaluated to determine optimal trajectories off the haul road.\\
\midrule

Control and Vehicle Interface
& Tracks the planned trajectory by regulating steering and acceleration, typically assuming Ackermann-steered vehicles.
& Must account for uneven terrain, wheel slip, steep slopes, and payload changes that alter mass and center of gravity, as well as articulated steering where applicable; hydraulic actuation introduces larger delay and stronger non-linearity than road-vehicle actuators. Additional machine-specific work equipment, such as the dump bed. \\

\bottomrule
\end{tabularx}
\end{table*}

\section{Technology Gaps in Autonomous Driving Between Public Roads and Construction Sites}

We identified key differences between public-road and off-road construction environments in each module of the Autoware stack, as summarized in Table~\ref{tab:technology_gap}. Challenges in construction environments are interdependent across these modules. Therefore, achieving construction autonomy requires system-wide integration rather than the isolated replacement of individual algorithms.

\section{Lessons Learned from Field Development of LiDAR-based Perception}
Among the technology gaps in Table~\ref{tab:technology_gap}, this section presents our LiDAR-based perception pipeline: LiDAR sensor $\rightarrow$ sensor driver $\rightarrow$ preprocessing filters $\rightarrow$ multi-LiDAR concatenation $\rightarrow$ postprocessing filters $\rightarrow$ ground segmentation $\rightarrow$ dust filtering $\rightarrow$ object detection, developed by extending the existing Autoware pipeline. Specifically, this section reports lessons learned from the LiDAR-based ground segmentation and dust filtering, as well as extrinsic sensor calibration. This work used Hesai AT128 LiDARs.

\subsection{Ground segmentation}

\begin{figure}[thpb]
    \centering
    \includegraphics[width=\columnwidth]{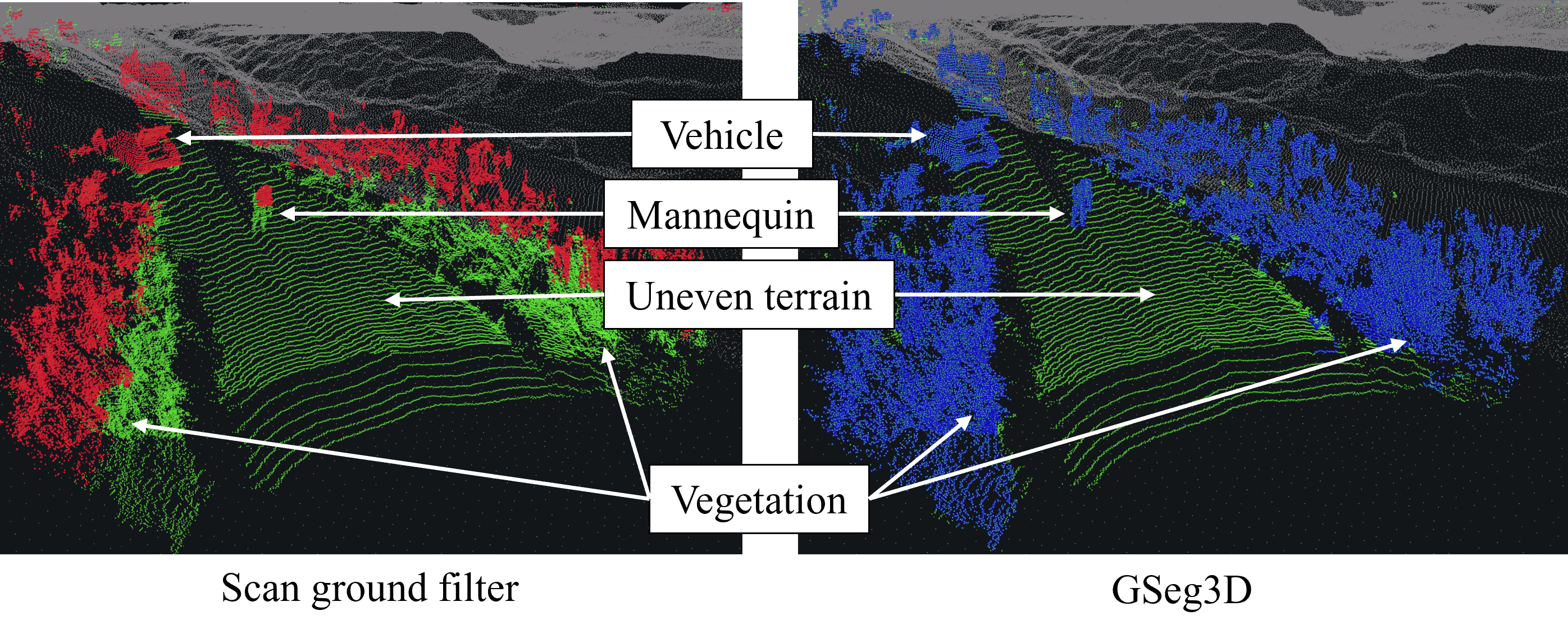}
    \caption{Comparison of LiDAR ground segmentation results obtained at an off-road environment using the scan ground filter (left) and GSeg3D (right). The raw point cloud is shown in green, while non-ground points identified by the scan ground filter and GSeg3D are highlighted in red and blue, respectively.}
    \label{fig:ground_segmentation}
\end{figure}

In Autoware, ground removal is commonly performed using \texttt{scan\_ground\_filter}~\cite{autoware_scan_ground_filter} to separate ground points from non-ground objects in LiDAR point clouds. The algorithm divides the point cloud into radial sectors and classifies points using slope and height thresholds. However, on uneven off-road terrain, one set of thresholds misclassifies uneven ground as obstacles, whereas the other set of thresholds incorrectly classifies small obstacles, such as rocks, as ground.

Therefore, ground segmentation algorithms designed for or adaptable to off-road environments are needed. Recent methods, such as \texttt{Patchwork++}~\cite{lee2022patchwork++} and \texttt{GSeg3D}~\cite{lodhi2025gseg3d}, have shown promising performance on public-road benchmarks. We compared \texttt{GSeg3D} with \texttt{scan\_ground\_filter}, as shown in Fig.~\ref{fig:ground_segmentation}. \texttt{GSeg3D} successfully classified the mannequin, vehicle, and vegetation as non-ground while preserving the uneven terrain as ground. In contrast, \texttt{scan\_ground\_filter} misclassified parts of these objects.

\begin{figure}[thpb]
    \centering
    \includegraphics[width=\columnwidth]{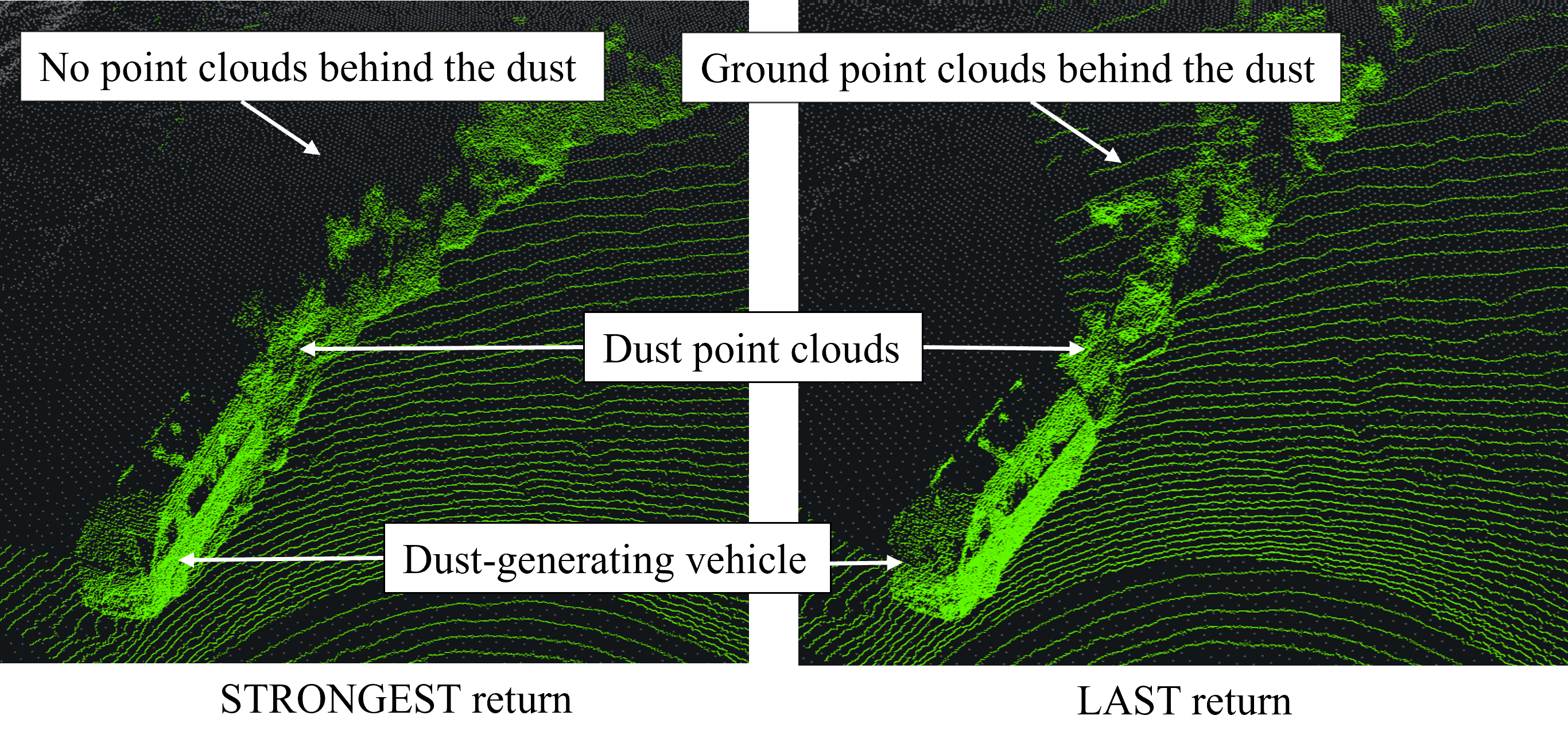}
    \caption{Comparison of LiDAR return modes under dusty conditions. The strongest return (left) is more affected by dust particles, whereas the last return (right) better preserves ground surfaces and objects behind the dust.}
    \label{fig:dust_return_type}
\end{figure}

For further development, traversability analysis is another important direction~\cite{lim2024similar}. Ground segmentation alone is insufficient because not all ground surfaces are safe for a vehicle to traverse. Traversability estimation should account for terrain properties, including slope, roughness, and material, while detecting positive obstacles, such as rocks, ruts, and berms, and negative obstacles, such as holes, trenches, and drop-offs. For example, a vehicle may be able to traverse a 30-cm-deep rut but should avoid a 30-cm-high rock. Such judgments therefore depend on vehicle geometry, dynamic state, payload, and operating conditions, which means traversability cannot be baked into a static map.

\subsection{Dust filters}
Airborne dust is a major challenge at construction sites. In our field deployments, it arose from various sources, including wind, tires traveling over dry soil, material spilling from a moving truck's dump bed, loading and dumping operations, and engine exhaust.

LiDAR return-mode selection is an important sensor configuration choice for operation in dusty environments. Many modern LiDARs support multiple return modes, including \texttt{STRONGEST}, \texttt{LAST}, and \texttt{FIRST}, which select the detected echo with the greatest received-signal strength, the farthest valid echo, and the nearest valid echo, respectively. Fig.~\ref{fig:dust_return_type} compares the \texttt{STRONGEST} and \texttt{LAST} return modes under dusty conditions. In the \texttt{STRONGEST} mode, dense nearby dust produces the strongest received echo, causing the dust cloud to dominate the point cloud. In contrast, the \texttt{LAST} mode better preserves the ground and other surfaces beyond the dust cloud and is therefore preferable in this setting. 

Beyond return-mode selection, dust returns can also be identified based on their characteristic properties. LiDAR returns from airborne dust typically exhibit low intensity, spatial dispersion, and limited temporal persistence~\cite{afzalaghaeinaeini2022design}. Although neighbor-search methods can leverage spatial and temporal information, they are generally computationally expensive for real-time operation. In contrast, simple intensity-based filtering has been shown to be effective~\cite{nakajima2024removing}.

\begin{figure}[thpb]
    \centering
    \includegraphics[width=\columnwidth]{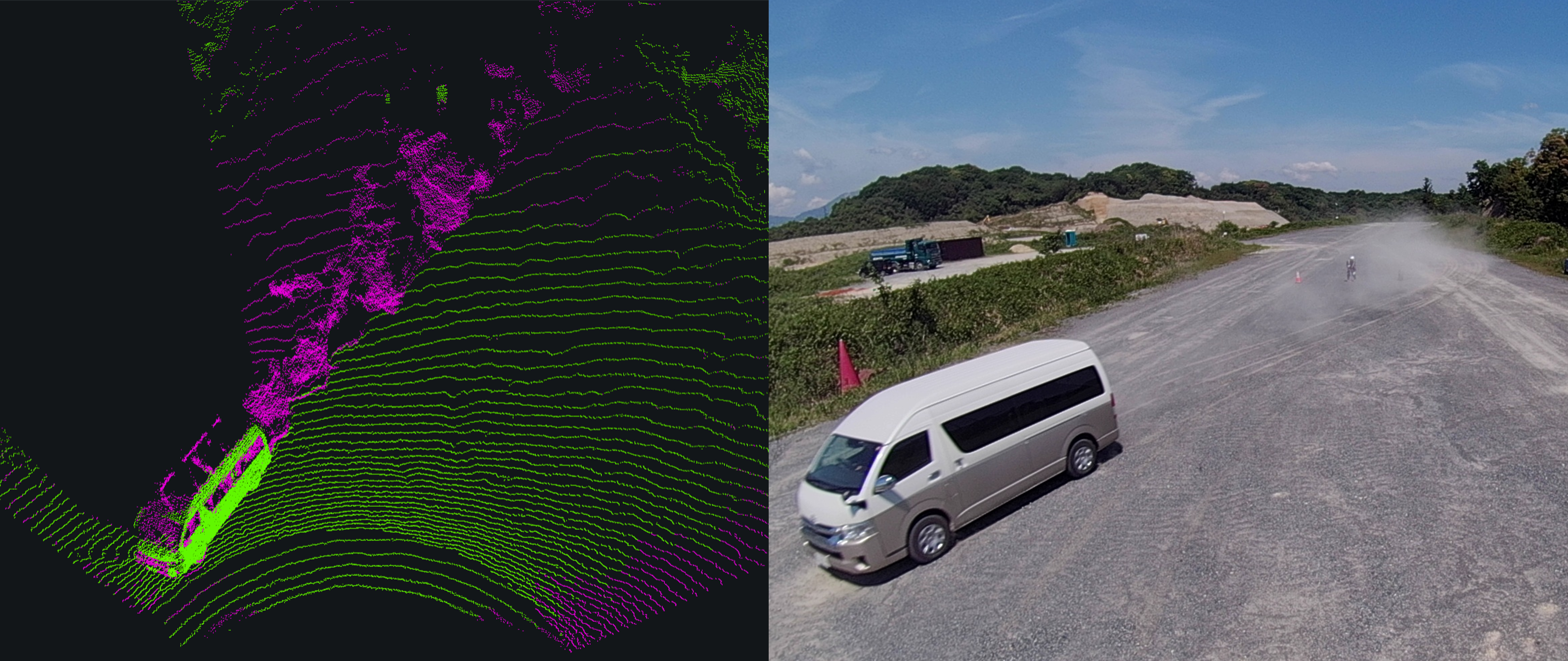}
    \caption{LiDAR point cloud (left) and camera image (right). In the LiDAR point cloud, LAST returns with intensity values of 1 or lower are shown in pink; others are shown in green. Dust returns and ground returns obscured by the dust are predominantly pink, whereas unobscured ground returns and the cones and mannequin behind the dust appear green.}
    \label{fig:dust_lidar_camera}
\end{figure}

This low-intensity characteristic of dust is evident in our experiment shown in Fig.~\ref{fig:dust_lidar_camera}, in which a vehicle generates a dense dust cloud. LiDAR points with intensity values of 1 or lower are highlighted in pink and closely correspond to returns from dust and ground obscured by the dust, while the intensities of cones and mannequins remain high. Based on this observation, an intensity-based filter can remove dust from non-ground LiDAR points identified by ground segmentation, thereby reducing false-positive object detections.

It should be noted that, depending on the sensor, the reported intensity may be a calibrated, range-independent quantity and may not directly correspond to the metric used for return selection. Consequently, a dust return in \texttt{STRONGEST} mode can still exhibit low intensity.

When multiple LiDARs are used, however, intensity-based filtering breaks down because intensity values are not directly comparable across LiDARs~\cite{viswanath2023off}. In our experience, differences in sensor architectures (e.g., spinning versus solid-state), as well as differences among manufacturers and among sensor models from the same manufacturer, result in distinct intensity ranges. Consequently, applying a single intensity threshold to a combined multi-LiDAR point cloud would require cross-sensor intensity normalization. Because such normalization is difficult in practical deployments, it is not adopted in this work.

Instead, we apply a sensor-specific low-intensity threshold in a preprocessing filter at the output of each LiDAR driver. LiDAR points below the low-intensity threshold are flagged separately for each LiDAR before the point clouds are concatenated, thereby eliminating the need for cross-sensor intensity normalization. These flagged points are retained during ground segmentation of the concatenated point cloud, since removing them beforehand would destabilize the ground estimation. After ground segmentation, non-ground points flagged as low-intensity are removed by the dust filter, while the remaining high-intensity non-ground points are passed to the subsequent object detection pipeline.

\subsection{Extrinsic sensor calibration}
Due to road-width limits and transportation regulations, large construction vehicles often cannot be transported fully assembled. Instead, they must be partially disassembled and reassembled at the construction site. Since sensors are often removed or repositioned during this process, extrinsic calibration must be performed or verified on-site.

Existing Autoware-based calibration workflows often assume the availability of flat ground and overlapping sensor fields of view containing distinctive environmental features. These assumptions may not hold at construction sites; suitable flat surfaces may be unavailable; the large dimensions of the vehicle and widely spaced sensors may yield limited overlap between sensor fields of view; and the environment may offer few distinctive features. Sensor calibration for large vehicles under such off-road conditions remains an active research topic~\cite{li2025online,wang2025joint}. Practical deployment therefore requires an on-site calibration method that can accommodate uneven terrain and limited sensor overlap.

\section{Roadmap toward End-to-End autonomous driving in construction}
\subsection{Addressing data scarcity}
Current learning-based perception modules in Autoware have been trained primarily on public-road datasets \cite{tanaka2025awml}. For example, a standard LiDAR-based CenterPoint model detects five object classes: cars, trucks, buses, bicycles, and pedestrians. Construction sites, however, contain a substantially different range of objects, including heavy construction machinery, service vehicles, workers wearing high-visibility clothing, traffic cones,  rocks, and berms.

Publicly available off-road datasets have largely been collected using small unmanned ground vehicles or passenger-sized vehicles, whose sensor viewpoints, sensing ranges, and occlusion patterns differ substantially from those of large construction machines. AutoMine~\cite{li2022automine} is among the most relevant datasets because it includes data collected from wide-body trucks and a mining haul truck. Nevertheless, construction-domain datasets remain scarce.

\subsection{End-to-end architecture}
Extending Autoware to construction environments should proceed incrementally. In the near term, the priority is to replace public-road assumptions with construction-specific capabilities while retaining Autoware's modular architecture. Meanwhile, on-site data collection can establish construction-specific datasets to address data scarcity and support the training of data-hungry E2E models. The longer-term objective is to integrate learned E2E models that jointly perform perception, reasoning, and trajectory generation. This direction aligns with Autoware's roadmap for gradually evolving its modular architecture toward E2E autonomy~\cite{autoware_e2e_report}.

Alpamayo–Autoware~\cite{alpamayo_autoware} provides an early example of this transition in public-road environments by integrating NVIDIA's Alpamayo~\cite{wang2025alpamayo} vision-language-action (VLA) model with Autoware's planning module. The model processes multi-camera observations and vehicle-motion history to generate an Autoware-compatible trajectory together with a textual explanation of its reasoning.

For construction applications, a VLA model could also accept task-level instructions such as positioning a haul truck for excavator loading, coordinating with a dozer in the dumping area, avoiding unstable terrain, and yielding to higher-priority vehicles.

\addtolength{\textheight}{-12cm}   




\section*{ACKNOWLEDGMENT}
We express our sincere gratitude to the following companies in Japan, listed in alphabetical order: EARTHBRAIN, Komatsu, and Taisei Corporation, for their contributions to this work. We also sincerely thank all TIER IV team members and collaborators for their contributions.

\bibliographystyle{IEEEtran}
\bibliography{references}

@inproceedings{kato2018autoware,
  title        = {Autoware on Board: Enabling Autonomous Vehicles with Embedded Systems},
  author       = {Kato, Shinpei and Tokunaga, Shota and Maruyama, Yuya and
                  Maeda, Seiya and Hirabayashi, Manato and Kitsukawa, Yuki and
                  Monrroy, Abraham and Ando, Tomohito and Fujii, Yusuke and
                  Azumi, Takuya},
  booktitle    = {Proceedings of the 2018 ACM/IEEE 9th International Conference
                  on Cyber-Physical Systems (ICCPS)},
  pages        = {287--296},
  year         = {2018},
  publisher    = {IEEE},
  doi          = {10.1109/ICCPS.2018.00035}
}

@article{uchiito2025robotics,
  title={Robotics Under Construction: Challenges on Job Sites},
  author={Uchiito, Haruki and Bhat, Akhilesh and Kusaka, Koji and Zhang, Xiaoya and Kinjo, Hiraku and Uehara, Honoka and Koyama, Motoki and Natsume, Shinji},
  journal={arXiv preprint arXiv:2506.19597},
  year={2025}
}

@misc{autoware_architecture,
  author       = {{Autoware Foundation}},
  title        = {{Autoware Architecture overview}},
  howpublished = {Available: \url{https://tier4.github.io/autoware-documentation/tier4-main/design/autoware-architecture/}},
  note         = {Accessed Aug. 1, 2026}
}

@misc{autoware_offroad,
  author       = {{Autoware Foundation}},
  title        = {{Autoware---Off-Road Autonomy Pilot}},
  howpublished = {Available: \url{https://github.com/autowarefoundation/autoware.off-road}},
  note         = {Accessed Aug. 1, 2026}
}

@misc{autoware_scan_ground_filter,
  author       = {{Autoware Foundation}},
  title        = {{Scan Ground Filter}},
  howpublished ={Available: \url{https://github.com/autowarefoundation/autoware_universe/blob/main/perception/autoware_ground_segmentation/docs/scan-ground-filter.md}},
  note         = {Accessed Aug. 1, 2026}
}

@inproceedings{lee2022patchwork++,
  title={Patchwork++: Fast and robust ground segmentation solving partial under-segmentation using 3D point cloud},
  author={Lee, Seungjae and Lim, Hyungtae and Myung, Hyun},
  booktitle={2022 IEEE/RSJ International Conference on Intelligent Robots and Systems (IROS)},
  pages={13276--13283},
  year={2022},
  organization={IEEE}
}

@inproceedings{lodhi2025gseg3d,
  title={GSeg3D: A High-Precision Grid-Based Algorithm for Safety-Critical Ground Segmentation in LiDAR Point Clouds},
  author={Lodhi, Muhammad Haider Khan and Hertzberg, Christoph},
  booktitle={2025 7th International Conference on Robotics and Computer Vision (ICRCV)},
  pages={119--126},
  year={2025},
  organization={IEEE}
}

@article{lim2024similar,
  title={Similar but different: A survey of ground segmentation and traversability estimation for terrestrial robots},
  author={Lim, Hyungtae and Oh, Minho and Lee, Seungjae and Ahn, Seunguk and Myung, Hyun},
  journal={International Journal of Control, Automation and Systems},
  volume={22},
  number={2},
  pages={347--359},
  year={2024},
  publisher={Springer}
}

@article{afzalaghaeinaeini2022design,
  title={Design of dust-filtering algorithms for lidar sensors using intensity and range information in off-road vehicles},
  author={Afzalaghaeinaeini, Ali and Seo, Jaho and Lee, Dongwook and Lee, Hanmin},
  journal={Sensors},
  volume={22},
  number={11},
  pages={4051},
  year={2022},
  publisher={MDPI}
}

@article{nakajima2024removing,
  title={Removing Snowfall Noise From Point Clouds Without Considering the Nearest Neighbors},
  author={Nakajima, Kensho and Nishio, Nobuhiko},
  journal={IEEE Sensors Journal},
  volume={25},
  number={3},
  pages={5212--5220},
  year={2024},
  publisher={IEEE}
}

@inproceedings{viswanath2023off,
  title={Off-road lidar intensity based semantic segmentation},
  author={Viswanath, Kasi and Jiang, Peng and Sujit, PB and Saripalli, Srikanth},
  booktitle={International Symposium on Experimental Robotics},
  pages={608--617},
  year={2023},
  organization={Springer}
}

@article{tanaka2025awml,
  title={AWML: An Open-Source ML-based Robotics Perception Framework to Deploy for ROS-based Autonomous Driving Software},
  author={Tanaka, Satoshi and Thapa, Samrat and Tan, Kok Seang and Szymko, Amadeusz and Kenzo, Lobos and Minoda, Koji and Tomie, Shintaro and Uetake, Kotaro and Zhang, Guolong and Yamashita, Isamu and others},
  journal={arXiv preprint arXiv:2506.00645},
  year={2025}
}

@article{li2025online,
  title={Online calibration of LiDAR and GPS/INS using multi-feature adaptive optimization in unstructured environments},
  author={Li, Huazhi and Yu, Guizhen and Wang, Zhangyu and Zhao, Fei and Chen, Peng},
  journal={IEEE Transactions on Instrumentation and Measurement},
  volume={74},
  pages={1--15},
  year={2025},
  publisher={IEEE}
}

@article{wang2025joint,
  title={Joint Optimization-based Targetless Extrinsic Calibration for Multiple LiDARs and GNSS-Aided INS of Ground Vehicles},
  author={Wang, Junhui and Qiao, Yan and Gao, Chao and Wu, Naiqi},
  journal={arXiv preprint arXiv:2507.08349},
  year={2025}
}

@inproceedings{li2022automine,
  title={AutoMine: An unmanned mine dataset},
  author={Li, Yuchen and Li, Zixuan and Teng, Siyu and Zhang, Yu and Zhou, Yuhang and Zhu, Yuchang and Cao, Dongpu and Tian, Bin and Ai, Yunfeng and Xuanyuan, Zhe and others},
  booktitle={2022 IEEE/CVF Conference on Computer Vision and Pattern Recognition (CVPR)},
  pages={21276--21285},
  year={2022},
  organization={IEEE}
}

@misc{autoware_e2e_report,
  author      = {{Autoware Foundation}},
  title       = {{Introducing Open-Source End-to-End Autonomous Driving}},
  howpublished = {Available: \url{https://autoware.org/wp-content/uploads/2025/10/Autoware-E2E-Report_Release.pdf}},
}

@misc{alpamayo_autoware,
  author       = {{Autoware Foundation}},
  title        = {{Alpamayo--Autoware: Alpamayo 1.5 ROS 2 Node}},
  howpublished = {Available: \url{https://github.com/autowarefoundation/alpamayo-autoware}},
  note         = {Accessed Aug. 1, 2026}
}

@article{wang2025alpamayo,
  title={Alpamayo-r1: Bridging reasoning and action prediction for generalizable autonomous driving in the long tail},
  author={Wang, Yan and Luo, Wenjie and Bai, Junjie and Cao, Yulong and Che, Tong and Chen, Ke and Chen, Yuxiao and Diamond, Jenna and Ding, Yifan and Ding, Wenhao and others},
  journal={arXiv preprint arXiv:2511.00088},
  year={2025}
}

\end{document}